(Updated on September 26, 2022)

**Title**
"Curse of rarity" for autonomous vehicles

**Authors**
Henry X. Liu[1,2,3,*] and Shuo Feng[2,*]
[1]Department of Civil and Environmental Engineering, University of Michigan, Ann Arbor, MI, USA
[2]University of Michigan Transportation Research Institute, Ann Arbor, MI, USA
[3]Mcity, University of Michigan, Ann Arbor, MI, USA
*Equally contributed and co-corresponding authors (henryliu@umich.edu, fshuo@umich.edu)

**Abstract**
In this paper, we reveal that the rarity of safety-critical events in high-dimensional driving environments is the root cause of the safety challenge for autonomous vehicle development. We formulate it as "curse of rarity" (CoR) because it occurs ubiquitously in various safety-critical systems such as medical diagnosis and aerospace systems.

**Main Text**
The dream of autonomous vehicles (AVs) is around a hundred years old[1]. In 1925, inventor Francis Houdina demonstrated a radio-controlled vehicle traveling through thick traffic in New York City streets, where the radio signals operated small electric motors that directed maneuvers of the vehicle[2]. Over the past 20 years, significant progress has been made on the development of AVs[3], particularly with the rapid development of computers and deep learning[4]. By 2015, a number of companies had announced that they would be mass producing AVs before 2020 (Ref.[5]). So far, however, the reality has not lived up to these expectations, and no SAE Level 4 (Ref.[6]) AVs are commercially available. The reasons for this are numerous. But above all, the safety performance of AVs is still significantly below that of human drivers.

To address the safety performance gap, the so-called "long-tail challenge" has been frequently discussed. With no clear definition, it usually refers to the problem that AVs should be able to handle seemingly endless low-probability safety-critical driving scenarios, even though millions of testing miles have been accumulated on public roads[7]. This applies to different components of AVs including perception, prediction, and planning, etc. However, most existing works mainly focus on a specific task such as the deep long-tailed learning for visual recognition[8]. To the best of the authors' knowledge, there is neither definition nor analysis of properties of the long-tail problems, which hinders the progress of addressing them.

In this paper, we systematically analyze the "long-tail challenge" and formulate it as the "curse of rarity" (CoR) for AVs. We reveal that the CoR - the rarity of safety-critical events and high dimensionality of driving environments - is the root cause of the "long-tail challenge". We discuss the CoR in various aspects of AV development including perception, prediction, and planning, as well as validation and verification. Based on these analyses and discussions, we discuss potential solutions to address the CoR in order to accelerate AV development and deployment. We hope that this paper can provide a better understanding of the safety challenges faced by the AV community, and a rigorous formulation of CoR can help accelerate the development and deployment of AVs.

We note that similar problems also exist for many other safety-critical applications, such as medical diagnosis and aerospace systems. For example, it is extremely challenging to detect a relatively rare disease with high precision and recall[9], due to the rarity of disease and the high dimensionality of diagnosis data. That is one critical reason why some of the IBM Watson's health products such as Watson for Oncology have not been as successful as expected[10]. Aerospace systems are another example safety-critical applications suffering from similar issues, such as verification and validation of an aircraft collision avoidance system[11].

**What is the "curse of rarity"?**

Before we start the discussion on "curse of rarity", let us discuss the "curse of dimensionality", which is well-known.

**Curse of dimensionality.** The term of "curse of dimensionality" was coined by Richard Bellman when considering problems in dynamic programming[12] and has been investigated in various domains[13], such as function approximation, numerical integration, optimization, combinatorics, sampling, machine learning, data mining, etc. The key concept of CoD is that when the dimensionality increases, the volume of the space increases so fast that the available data become sparse. Therefore, in order to obtain a reliable result, the amount of data needed often grows exponentially with the dimensionality. For example, if we need to optimize, approximate, or integrate a function of $d$ variables and we know only that it is Lipschitz, we need order of $(1/\epsilon)^d$ evaluations on a grid in order to obtain the results within error $\epsilon$.

In real-world applications, the dimensionality of problems usually grows quickly with the increase of problem complexity. A dataset in domains such as healthcare and genomics can easily include thousands or millions of features. For example, the input vector of an image could include $224 \times 224 \approx 5 \times 10^4$ pixels, which is high dimensional. Moreover, variables that are needed to describe spatiotemporally complex problems are usually high dimensional. For example, the board game Go has a state space of $10^{360}$ (Ref.[14]) and the semiconductor chip design could have a state space on the order of $10^{2500}$ (Ref.[15]). Autonomous driving is also a high-dimensional problem because the complexity of driving environments can be impacted by different weather conditions, different roadway infrastructures, different road users and each with different behavior, etc. For example, an AV driving in a city environment for an extended period of time could interact with many vehicles and other road users (e.g., pedestrians) on multiple road facilities (e.g., highway, intersection, roundabout, etc.) under different weather and lighting conditions. Therefore, the models to define such complex driving environments are high dimensional.

To address this challenge, significant progress has been made during the past decades, particularly with the emergence of deep learning[4]. One representative approach is to convert high-dimensional data to low-dimensional codes by training a multilayer neural network with a small central layer to reconstruct high-dimensional input vectors[16]. By utilizing gradient descent for fine-tuning the weights in neural networks, the dimensionality of data can be significantly reduced, which can help overcome the CoD and enable the numerous pattern recognition algorithms. Another representative approach for high-dimensional problems is to learn a decision-making policy through deep reinforcement learning (DRL)[17]. It converts the optimization problem in a high-dimensional variable space to a parameter space of neural networks utilizing techniques such as policy gradient theory, bootstrapping, and Monte Carlo Tree Search, which can help overcome the CoD. This approach has reached superhuman performance in many spatiotemporally complex domains such as chip design[15] and various games[14].

Although the deep learning approaches have achieved rapid progress for solving problems with CoD, their applications in safety-critical systems remain challenging, such as AVs, medical diagnosis, and aerospace systems. The deadly consequences of safety-critical events demand maximum safety performance guarantee, while the rarity of safety-critical events makes it difficult for learning. This leads to the discussion on the "curse of rarity" in the next subsection.

**Curse of rarity.** In this study, we argue that the rarity of safety-critical events causes a fundamental challenge for AVs, and it is completely different from the CoD. As it occurs ubiquitously in various domains, we call it the curse of rarity (CoR). The basic concept of CoR is that the occurrence probability for the events of interest is so rare that most available data contain little information regarding the rare events. Therefore, in order to obtain sufficient information of rare events, the amount of data needed often grows dramatically with the rarity of events. More importantly, due to the rarity of events of interests, it will be difficult for us to optimize the system performance regarding the rare events by using the information obtained through observation and analysis of the system's behavior. This is because, with the decrease in the occurrence probability of events, the policy gradient estimation will have larger variance and smaller signal-to-noise ratio, which may mislead the optimization process. In other words, due to the rarity of events, it is difficult to evaluate the performance sensitivities in the policy space, which hinders the applicability of deep learning techniques for safety-critical systems.

To further understand the CoR, let us consider a general deep learning problem that can be formulated as an optimization problem:
$$\max_{\theta} \mathbb{E}_P[f_\theta(x)], \tag{1}$$
where $\theta$ denotes the parameters of a neural network, $x \in \Omega$ denotes the training data with an underlying distribution $P$, and $f_\theta(x)$ denotes the objective function given the neural network $\theta$ and training data $x$. (For supervised learning, $x$ could denote the labeled data and $f_\theta(x)$ could denote the loss function, while for reinforcement learning, $x$ could denote the environment data and $f_\theta(x)$ could denote the reward function.) To optimize the objective function, the key to the deep learning problem is to estimate the policy gradient of the neural network at each training iteration as
$$\mu \stackrel{\text{def}}{=} \nabla_\theta \mathbb{E}_P[f_\theta(x)] \approx \frac{1}{n} \sum_{i=1}^{n} \nabla_\theta f_\theta(x_i), \tag{2}$$
where $n$ denotes the number of training data samples at each iteration and the approximation is obtained using the Monte Carlo method[18]. To simplify the notations, we denote $X \stackrel{\text{def}}{=} \nabla_\theta f_\theta(x): \Omega \to \mathbb{R}$ as a random variable, so the gradient estimation in Eq. (2) can be denoted as
$$\mu_1 \stackrel{\text{def}}{=} \frac{1}{n} \sum_{i=1}^{n} X_i. \tag{3}$$
It can be easily proved that $\mu_1$ is an unbiased estimation of $\mu$, i.e., $\mathbb{E}_P(\mu_1) = \mu$, and the variance of the estimator is
$$Var_P(\mu_1) = \sigma_{\mu_1}^2/n, \sigma_{\mu_1}^2 = Var_P[X]. \tag{4}$$

Now let's focus on a special set of deep learning problems where only a very small portion of training data (critical data) can contribute effectively to the gradient estimation, while a vast majority of training data (normal data) contributes little. For AVs, most scenarios are not safety-critical and training the neural network in these scenarios cannot effectively improve the safety performance. Safety-critical events, for example near-miss events or crashes, are rare, but contain much valuable information for AV safety training. Similar problems exist in other safety-critical systems such as medical diagnosis and aerospace systems, because of the rarity of the events of interest.

To be more specific, we can define normal events $A \subset \Omega$ and critical but rare events $B \subset \Omega$, and their corresponding indicator functions $\mathbb{I}_A$ and $\mathbb{I}_B$ with the following properties:
$$A \cap B = \emptyset, A \cup B = \Omega, \mathbb{E}_P[X \cdot \mathbb{I}_A(X)] = 0, \mathbb{E}_P[X \cdot \mathbb{I}_B(X)] \neq 0. \tag{5}$$
Therefore, due to the properties of Monte Carlo method for rare events[18], the policy gradient estimator $\mu_1$ will suffer from the large estimation variance $Var_P(\mu_1)$, which would mislead the learning process.

As the expectation of the samples associated with the event $A$ is zero ($\mathbb{E}_P[X \cdot \mathbb{I}_A(X)] = 0$), another way to estimate the gradient is to only utilize the samples associated with the event B as
$$\mu_2 \stackrel{\text{def}}{=} \frac{1}{n} \sum_{i=1}^{n} (X_i \cdot \mathbb{I}_B(X_i)), \tag{6}$$
with the variance as

$$Var_P(\mu_2) = \sigma_{\mu_2}^2/n, \sigma_{\mu_2}^2 = Var_P[X \cdot \mathbb{I}_B(X)]. \tag{7}$$

Then, we have the following theorem, and the proof can be found in the Appendix.

**Theorem 1**:
The estimator $\mu_2$ has the following properties:
(1) $\mathbb{E}_P(\mu_2) = \mathbb{E}_P(\mu_1) = \mu$;
(2) $\sigma_{\mu_1}^2 \geq \sigma_{\mu_2}^2$; and
(3) $\sigma_{\mu_1}^2 \geq \rho_B^{-1} \sigma_{\mu_2}^2$, with the assumption
$$\mathbb{E}_P(X^2 \cdot \mathbb{I}_B(X)) = \mathbb{E}_P(X^2) \cdot \mathbb{E}_P(\mathbb{I}_B(X)), \tag{8}$$
where $\rho_B = \mathbb{E}_P(\mathbb{I}_B(X)) \in [0,1]$ is the expected proportion of the event $B$ in all samples with the sampling distribution $P$.

**Remark 1**. The first property indicates that $\mu_2$ is also an unbiased estimation, which is reasonable as the mean value of the samples associated with the event $A$ is zero. The second property indicates that the estimator $\mu_2$ is at least not worse than the estimator $\mu_1$ regarding the estimation variance. To quantify the variance reduction, we further introduce the assumption in Eq. (8), where $X^2$ is independent on the event $B$. Then, the third property indicates that the estimator $\mu_2$ could reduce the variance of the estimator $\mu_1$ by the expected proportion of the event $B$ in all samples.

**Remark 2**. Since the events in $B$ are rare, $\rho_B$ is a very small value (e.g., $10^{-6}$). Moreover, for deep learning approaches, the random variable $X \stackrel{\text{def}}{=} \nabla_\theta f_\theta(x)$ is usually dependent on neural networks with random initialization. Therefore, $X$ could have a stationary uncertainty that is independent of the events $A$ and $B$, particularly at the beginning of the learning process, which satisfies the assumption in Eq. (8). According to Theorem 1, as $\rho_B$ could be very small, $\sigma_{\mu_1}^2 \geq \rho_B^{-1} \sigma_{\mu_2}^2$ could be an extremely high variance, which would require a very large sample size for the estimation of $\mu$. Moreover, as $\mu$ is usually very small, the signal-to-noise ratio ($\mu^2/\sigma_{\mu_1}^2$) is also rare, which makes the problem even worse. Such a high variance and rare signal-to-noise ratio could bring difficulties to many optimization and learning techniques, which results in the CoR. An illustration can be found in Figure 1.

**Remark 3**. The estimator $\mu_2$ looks like a natural way to address the above issues. Intuitively, if we can train the model with set $B$ instead of $\Omega$, then the training process will be much more effective, as $\sigma_{\mu_2}^2$ is much smaller than $\sigma_{\mu_1}^2$. However, how to define and identify the rare event is challenging, which depends on the problem-specific objective functions. We note that simply including crash events in $B$ may not work for AV problems. It is because that the rare event $B$ need to be defined according to the random variable $X \stackrel{\text{def}}{=} \nabla_\theta f_\theta(x)$ such that $\mathbb{E}_P(X \cdot \mathbb{I}_B(X))$ contains most of the information as shown in Eq. (5). Taking the AV training problem as an example, the objective function could relate to the crash rate of the AV, so the event $B$ should not only contain the crash data, but also contain the near-miss data that the AV avoids crashes successfully. Only with both types of data, the policy gradient $\nabla_\theta f_\theta(x)$ can be calculated accurately considering both penalties and rewards. Moreover, the definition of the rare event $B$ may also need to consider the temporal relations. For example, even for a crash trajectory, a portion of the trajectory may not be safety-critical until some adversarial maneuvers emerge. As these non-safety-critical states may contain no information of $\nabla_\theta f_\theta(x)$, they should not be considered as the event $B$. We believe that further investigations are deserved in this direction.

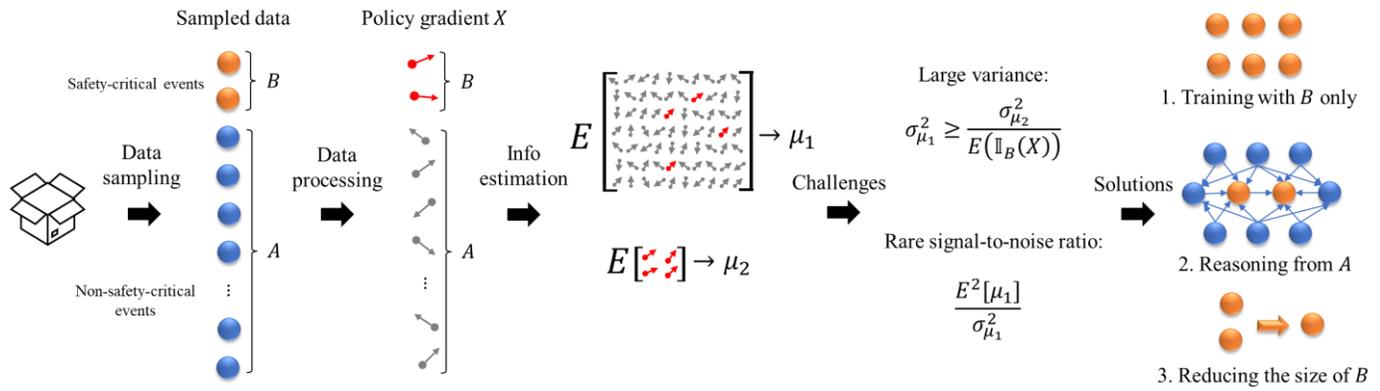

Figure 1. Illustration of the curse of rarity and potential solutions.

For AV development, both CoR and CoD problems exist —the rarity of safety-critical events in high-dimensional driving environments—are the root causes of various safety challenges in the development and deployment of AVs. On one hand, the CoD obstructs the applications of existing rare event simulation approaches for solving the CoR. For example, it has been shown that the variance of importance sampling approaches could grow exponentially with the dimensionality of variables[19], which is why most existing importance sampling-based AV testing methods can only be applied to short scenario segments with limited dynamic objects[20]. On the other hand, the CoR hinders the applications of deep learning approaches for solving the CoD. For example, the rarity of events can cause severe imbalanced data issues with a much greater imbalanced ratio than typical imbalance problems[21]. It can also cause the sparse reward issues and high variance issues of the policy gradient estimation, which hinder the applicability of DRL approaches.

The high-performance requirements (human or superhuman level) for AVs make the CoR even worse. Some commentators have claimed that AVs need to be safer than human drivers with the factor ranging from 2 to 100, so the fatality rate per mile of AVs may need to be smaller than $10^{-10}$. Improving AVs' safety performance with such a rarity of events is extremely challenging due to the CoR. As illustrated in Figure 2, with the decrease of exposure frequency, the development cost for handling the safety-critical events increases significantly. For example, it would take on average $10^8$ miles to collect a fatality event of AVs at the level of human drivers[22]. As the black curve looks like a long tail, the problem is also referred as to the "long-tail problem". The CoR occurs not only for verification and validation of AVs but also for perception, prediction, and planning, as discussed in the next section.

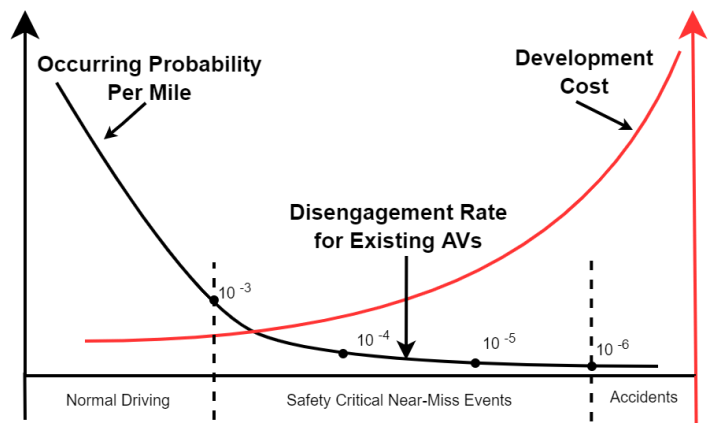

Figure 2. Illustration of the curse of rarity for AVs. The disengagement rate for existing AVs is around $2.0 \times 10^{-5}$ per mile, according to the 2021 Disengagement Report from California[23].

**What challenges it brings for autonomous vehicles?**

In this section, we will discuss the CoR in various aspects of AVs, including perception, prediction, planning, and validation and verification, etc.

**Perception**. Deep learning techniques have been used extensively in perception tasks. As training samples typically exhibit a long-tailed and open-ended distribution, the CoR leads to various challenges for perception tasks such as imbalance classification, few-short learning, and open-set recognition[8,24], particularly considering the high-performance requirements of AVs as discussed above. Taking the imbalance classification tasks as example, AVs require high precision and higher recall performances, as missing any safety-critical scenario could be fatal while too many false alarms could make AVs over-conservative. Moreover, most existing studies focus on the image level with a limited imbalance ratio (e.g., $10^2$)[21,24]. However, as safety-critical scenarios are usually spatiotemporally complex, perception algorithms may need to consider both temporal and spatial dimensions, which further complicates the issue and makes the imbalanced ratio much larger (e.g., $10^6$)[22].

**Behavior modeling and prediction**. To reach human-level or superhuman safety performance for AVs, it is critical to model and predict safety-critical behaviors of other road users with high statistical accuracy. Overestimating the probability of safety-critical behaviors could lead to over-conservative decision making, while underestimating the probability could lead to unexpected failures. This is particularly challenging considering the complexity of real-world driving environments and the rare probability (e.g., $10^{-6}$) of safety-critical behaviors, where a small absolute error could be fatal. To the best of the authors' knowledge, although advanced deep learning techniques have been widely applied recently, their performances are still far from the high-performance requirements, where the CoR is a major barrier. These performance gaps limit the performance of downstream tasks including decision making, safety metric development, testing and training, etc. For example, the behavior modeling errors will cause simulation-to-reality gaps, hindering the effectiveness of simulation-based AV testing and training[25].

**Decision making**. The most critical challenge of decision-making is to guarantee safety under all driving circumstances. Deep learning techniques have been adopted such as deep imitation learning[26] and DRL[27], which perform well in common scenarios. For safety-critical scenarios, however, the data is severely limited or even lacking, so the imitation learning approach may fail. More importantly, even given sufficient data, it is still hard for a deep learning model to learn due to the CoR. Therefore, the DRL approach will also suffer from the large variance and sparse reward issues during the training process. Another alternative approach aims to guarantee safety of decision-making using formal methods based on a set of assumptions. Due to the CoD, however, it is usually challenging to design and verify the assumptions that could be scenario dependent.

**Verification and validation**. Due to the rarity of safety-critical events, it is well-known that hundreds of millions of miles and sometimes hundreds of billions of miles would be required to validate the safety performance of AVs at the level of human drivers[22], which is intolerably inefficient. To address this challenge, many approaches have been developed, for example, scenario-based approaches[20] aim to test AVs in purposely generated scenarios using traditional rare event simulation methods. Due to the CoD, however, most existing approaches can only handle short scenario segments with limited background road users, which cannot represent the full complexity and variability of real-world driving environments[25]. To overcome this limitation, integrating existing testing approaches with deep learning techniques has great potential, where the CoR needs to be addressed.

**What are the potential solutions toward solving the "curse of rarity"?**

To date, the progress to address the CoR problem is limited as the underlying problem formulation does not exist until this study. Based on the analyses and discussions above, we identify three potential approaches for solving the CoR problem, as illustrated in Figure 1. We hope these discussions will provide directions for future development.

**Approach #1: Training with $B$**. The first approach is to utilize data associated with rare events $B$ only, as indicated by the estimator $\mu_2$ in Eq. (6), which could significantly reduce the estimation variance. As discussed in Remark 3, the key challenge is how to define, identify, and utilize the rare events for specific problems, particularly considering the spatiotemporal complexity of autonomous driving tasks. Moreover, different tasks usually utilize different techniques, such as supervised learning and DRL, with different objective functions, which could bring different requirements and constraints to training with the rare events. One example of this approach is the "shadow mode testing"[28], where Tesla proposed to identify the rare events of interest by comparing the human driving behaviors and autonomous driving behaviors, but no details are given in the literature.

**Approach #2: Reasoning from $A$**. The second approach is to reason the information of rare events $B$ from normal events $A$. Intuitively, as human drivers could drive vehicles with limited driving experience (for example, the training time for human drivers is usually less than 100 hours), next-generation AI agents might be able to overcome the CoR without relying on large-scale data. This requires an AI agent have both bottom-up reasoning (sensing data-driven) and top-down reasoning (cognition expectation-driven) capabilities[29], which would fill in the gap for the information not included in the data. Some recent studies aim to utilize physics knowledge to guide the training process with small data, such as physic-informed machine learning, but how to apply it to safety-critical systems remains an open problem.

**Approach #3: Reducing the size of $B$:** The third approach is to mitigate the impact of CoR by reducing the set $B$ so that the occurrence probability of safety-critical events becomes acceptable for human society. Infrastructure-assisted automated driving is one promising approach in this direction. Infrastructure-based sensors and vehicle-to-everything communication may help overcome the limitations of on-board sensors (due to range and occlusion) to reduce the occurring probability of safety-critical situations[30]. Cooperative perception and decision-making may help overcome the limitations of AVs in rare events.

All the above approaches face significant challenges, but their combinations may have great potential for solving the CoR issue and accelerating the large-scale deployment of AVs. This will be another direction worth further investigation.

**Conclusions**

In this study, we systematically discussed the curse of rarity (CoR) for AVs. We found that the CoR problem is the underlying cause for various critical challenges of AVs, including perception, prediction, decision making, and verification and validation, etc. In particularly, the CoR problem hinders the application of deep learning methods. We also propose potential approaches to address these challenges. It is our hope that this paper will provide a better understanding of the challenges faced by the industry and attract more attention from the community for further investigation.

**Appendix**
   (1) **Proof of $\mathbb{E}_P(\mu_2) = \mathbb{E}_P(\mu_1) = \mu$:**

$$\mathbb{E}_P(\mu_2) = \mathbb{E}_P\left(\frac{1}{n}\sum_{i=1}^{n}\left(X_i \cdot \mathbb{I}_B(X_i)\right)\right) = \mathbb{E}_P\left(X_i \cdot \mathbb{I}_B(X_i)\right) = \mu = \mathbb{E}_P(\mu_1).$$

*End of proof.*

**(2) Proof of $\sigma^2_{\mu_2} \leq \sigma^2_{\mu_1}$:**

$$\sigma^2_{\mu_2} = Var_P[X \cdot \mathbb{I}_B(X)] = \mathbb{E}_P[X^2 \cdot \mathbb{I}_B(X)] - \mathbb{E}_P^2[X \cdot \mathbb{I}_B(X)] = \mathbb{E}_P[X^2 \cdot \mathbb{I}_B(X)] - \mathbb{E}_P^2(X)$$
$$\leq \mathbb{E}_P[X^2 \cdot \mathbb{I}_B(X)] + \mathbb{E}_P[X^2 \cdot \mathbb{I}_A(X)] - \mathbb{E}_P^2(\mu_1) = \mathbb{E}_P[X^2] - \mathbb{E}_P^2(X) = \sigma^2_{\mu_1}.$$

*End of proof.*

**(3) Proof of $\sigma^2_{\mu_2} \leq \rho_B \sigma^2_{\mu_1}$, with the assumption $\mathbb{E}_P(X^2 \cdot \mathbb{I}_B(X)) = \mathbb{E}_P(X^2) \cdot \mathbb{E}_P(\mathbb{I}_B(X))$:**

$$\sigma^2_{\mu_2} = \mathbb{E}_P[X^2 \cdot \mathbb{I}_B(X)] - \mathbb{E}_P^2(X) = \mathbb{E}_P(X^2) \cdot \mathbb{E}_P(\mathbb{I}_B(X)) - \mathbb{E}_P^2(X)$$
$$\leq \mathbb{E}_P(X^2) \cdot \mathbb{E}_P(\mathbb{I}_B(X)) - \mathbb{E}_P^2(X) \cdot \mathbb{E}_P(\mathbb{I}_B(X)) = \mathbb{E}_P(\mathbb{I}_B(X)) \cdot [\mathbb{E}_P(X^2) - \mathbb{E}_P^2(X)] = \rho_B \sigma^2_{\mu_1}.$$

*End of proof.*

**Acknowledgements**
This research was partially funded by the U.S. Department of Transportation (USDOT) Region 5 University Transportation Center: Center for Connected and Automated Transportation (CCAT) of the University of Michigan. Any opinions, findings, and conclusions or recommendations expressed in this material are those of the authors and do not necessarily reflect the official policy or position of the Department of Transportation or the U.S. government.

**References**
1. Ronan Glon and Stephen Edelstein (2020).
   https://www.digitaltrends.com/cars/history-of-self-driving-cars-milestones
2. Time Magazine. Aug 10, 1925. "Science: Radio Auto". Retrieved 29 September 2013.
3. Jacobstein, N., 2019. Autonomous vehicles: an imperfect path to saving millions of lives. *Science Robotics*, **4**(28), DOI: 10.1126/scirobotics.aaw8703.
4. LeCun, Y., Bengio, Y., & Hinton, G. (2015) Deep learning. *Nature* **521**, 436-444.
5. Safe driving cars. *Nature Machine Intelligence* **4**, 95–96 (2022). https://doi.org/10.1038/s42256-022-00456-w.
6. Society of Automotive Engineers (2021) Taxonomy and Definitions for Terms Related to Driving Automation Systems for On-Road Motor Vehicles.
   https://www.sae.org/standards/content/j3016_202104/
7. Drago Anguelov. (2019) Taming the long tail of autonomous driving challenges.
   https://www.youtube.com/watch?v=Q0nGo2-y0xY&t
8. Liu, Z., Miao, Z., Zhan, X., Wang, J., Gong, B. and Yu, S.X., 2019. Large-scale long-tailed recognition in an open world. *In Proceedings of the IEEE/CVF Conference on Computer Vision and Pattern Recognition* (pp. 2537-2546).
9. Wiens, J., Saria, S., Sendak, M., Ghassemi, M., Liu, V.X., Doshi-Velez, F., Jung, K., Heller, K., Kale, D., Saeed, M. and Ossorio, P.N. (2019) Do no harm: a roadmap for responsible machine learning for health care. *Nature medicine*, **25**(9), pp.1337-1340.
10. Strickland, E. (2019). IBM Watson, heal thyself: How IBM overpromised and underdelivered on AI health care. *IEEE Spectrum*, **56**(4), 24-31.
11. Lee, R., Kochenderfer, M.J., Mengshoel, O.J., Brat, G.P. and Owen, M.P. (2015) Adaptive stress testing of airborne collision avoidance systems. *In 2015 IEEE/AIAA 34th Digital Avionics Systems Conference (DASC)*. DOI: 10.1109/DASC.2015.7311450
12. Bellman, Richard (1961) Adaptive Control Processes: A Guided Tour. Princeton University Press.
13. Donoho, D. L. (2000) High-dimensional data analysis: The curses and blessings of dimensionality. *AMS Math Challenges Lecture* **1**, 32.

14. Silver, D., Schrittwieser, J., Simonyan, K., Antonoglou, I., Huang, A., Guez, A., ... & Hassabis, D. (2017) Mastering the game of go without human knowledge. *Nature* **550**, 354-359.
15. Mirhoseini, A., Goldie, A., Yazgan, M., Jiang, J. W., Songhori, E., Wang, S., ... & Dean, J. (2021) A graph placement methodology for fast chip design. *Nature* **594**, 207-212.
16. Hinton, G. E., & Salakhutdinov, R. R. (2006) Reducing the dimensionality of data with neural networks. *Science* **313**, 504-507.
17. Sutton, R. S., & Barto, A. G. Reinforcement Learning: An Introduction. MIT press (2018).
18. Owen, A. B. (2013) Monte Carlo Theory, Methods and Examples. https://statweb.stanford.edu/~owen/mc/
19. Au, S. K. & Beck, J. L.(2003) Important sampling in high dimensions. *Struct. Saf.* **25**,139–163.
20. Riedmaier, S., Ponn, T., Ludwig, D., Schick, B. and Diermeyer, F., 2020. Survey on scenario-based safety assessment of automated vehicles. *IEEE Access*, **8**, pp.87456-87477.
21. Megahed, F. M., Chen, Y. J., Megahed, A., Ong, Y., Altman, N., & Krzywinski, M. (2021) The class imbalance problem. *Nature Methods* **18**, 1270-1272.
22. Kalra, N., & Paddock, S. M. (2016) Driving to safety: How many miles of driving would it take to demonstrate autonomous vehicle reliability?. *Transp. Res. A: Policy Pract.* **94**, 182-193.
23. California Department of Motor Vehicles (2022) 2021 Disengagement Reports https://www.dmv.ca.gov/portal/vehicle-industry-services/autonomous-vehicles/disengagement-reports/
24. Johnson, J. M., & Khoshgoftaar, T. M. (2019). Survey on deep learning with class imbalance. *Journal of Big Data*, **6**(1), 1-54.
25. Feng, S., Yan, X., Sun, H., Feng, Y., & Liu, H. X. (2021) Intelligent driving intelligence test for autonomous vehicles with naturalistic and adversarial environment. *Nature Communications* **12**, 1-14.
26. Le Mero, L., Yi, D., Dianati, M., & Mouzakitis, A. (2022). A Survey on Imitation Learning Techniques for End-to-End Autonomous Vehicles. *IEEE Transactions on Intelligent Transportation Systems*. DOI: 10.1109/TITS.2022.3144867.
27. Kiran, B.R., Sobh, I., Talpaert, V., Mannion P., Sallab A., Yogamani S., and Perez. P. (2022) Deep Reinforcement Learning for Autonomous Driving: A Survey*, IEEE Transactions on Intelligent Transportation Systems*. DOI: 10.1109/TITS.2021.3054625
28. Karpathy, A., Tesla Inc. (2021) System and method for obtaining training data. U.S. Patent Application 17/250,825.
29. Cummings, M. L. (2021). Rethinking the Maturity of Artificial Intelligence in Safety-Critical Settings. *AI Mag.*, **42**(1), 6-15.
30. Bai, Z., Wu, G., Qi, X., Liu, Y., Oguchi, K. and Barth, M.J. (2022) Infrastructure-Based Object Detection and Tracking for Cooperative Driving Automation: A Survey. https://arxiv.org/abs/2201.11871.